# Scale-Dependent Semantic Dynamics Revealed by Allan Deviation

## Debayan Dasgupta[1,2]

1. Theranautilus Pvt. Ltd, Bangalore
2. Center for Nanoscience and Engineering, Indian Institute of Science, Bangalore

Contact: debayan@theranautilus.com

## Abstract

While language progresses through a sequence of semantic states, the underlying dynamics of this progression remain elusive. Here, we treat the semantic progression of written text as a stochastic trajectory in a high-dimensional state space. We utilize Allan deviation, a tool from precision metrology, to analyze the stability of meaning by treating ordered sentence embeddings as a displacement signal. Our analysis reveals two distinct dynamical regimes: short-time power-law scaling, which differentiates creative literature from technical texts, and a long-time crossover to a stability-limited noise floor. We find that while large language models successfully mimic the local scaling statistics of human text, they exhibit a systematic reduction in their stability horizon. These results establish semantic coherence as a measurable physical property, offering a framework to differentiate the nuanced dynamics of human cognition from the patterns generated by algorithmic models.

## I. Introduction

Language can be treated as a complex emergent system in which meaning unfolds sequentially, with each sentence conditioning those that follow. Classical results such as Zipf's law, long-range correlations, and small-world organization demonstrate that language is not random, but they do not address how semantics evolve as a text progress[1], [2], [3], [4]. If a narrative is viewed as a time-ordered trajectory through a semantic state space, a fundamental question arises: does its evolution resemble a stochastic random walk, or does it exhibit long-range correlations analogous to those found in critical or driven systems?

Previous studies have established the presence of long-range correlations[5], [6], [7], [8], [9] in written language using random-walk mappings, power spectra, and detrended fluctuation analysis[1], [3], [4]. These approaches successfully identify scale-free structure but do not explicitly probe temporal ordering, and therefore cannot directly characterize the dynamics of semantic progression through text[2], [10], [11]. As a result, they do not provide a direct means to identify characteristic scales at which qualitative changes in semantic organization occur.

We address this question by mapping the sentences in a text corpus onto a continuous metric space. We rely on the distributional hypothesis, which posits that meaning is a relational property derived from linguistic context rather than an intrinsic attribute of individual symbols. While early implementations were inherently static, the advent of transformer-based architectures[12] and contextual embedding models such as BERT enables sentence-level representations[13], [14] that evolve with local context, capturing nuances in semantic interpretation like for example, distinguishing a river "bank" from a financial one. From a physics perspective, this representation allows semantic progression to be treated as a dynamically varying temporal signal, allowing us to probe whether its fluctuations exhibit scale-dependent correlations.

Here we formulate semantic progression as an ordered sequences of sentence embeddings and treat sentence index as a discrete time variable. Distances between sentences, mapped to a fixed-dimensional embedding vector and its successive sentences, define incremental semantic displacements. Accumulating these displacements yields a one-dimensional signal that preserves temporal ordering while discarding absolute semantic coordinates. We probe the scale-dependent fluctuations of this signal using Allan deviation[15], [16], [17], a variance estimator developed in precision metrology to distinguish short-time variability from long-range drift in sequential data.

## II. Results

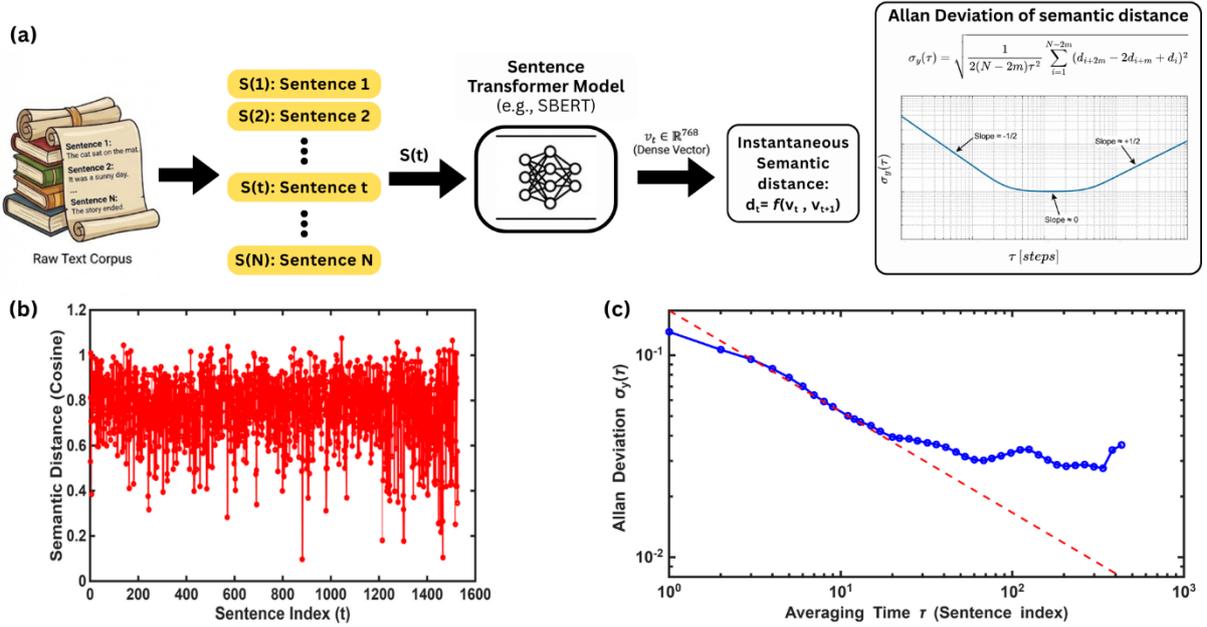

**Figure 1. Construction of the semantic signal and Allan deviation analysis.**
(a) Schematic of the analysis pipeline is shown. An ordered text corpus is segmented into sentences $S(t)$, which are mapped to fixed-dimensional embedding vectors $v_t$ using a sentence-transformer model. Successive sentences define an instantaneous semantic distance $d_t = f(v_t, v_{t+1})$, yielding a one-dimensional semantic signal indexed by sentence number. Allan deviation is then computed from the cumulative semantic displacement to probe scale-dependent stability. (b) Example time series of instantaneous semantic distance as a function of sentence index. (c) Allan deviation $\sigma_y(\tau)$ of the corresponding semantic phase, illustrating short-time power-law scaling and a crossover at larger averaging scales. The dashed line represents a slope of -0.5 which is the scaling for a memoryless random walk.

Figure 1(a) illustrates the construction of the semantic signal of a text corpus. Each sentence is mapped to a fixed-dimensional embedding vector, and successive sentences define an incremental semantic displacement $d_t$, computed from the cosine distance between their embedding vectors, an example of which is shown in Figure 1(b). The cumulative semantic phase, $\phi(t) = \sum_{i \leq t} d_i$, defines a one-dimensional stochastic signal indexed by sentence number. This representation preserves temporal ordering while discarding absolute semantic coordinates, allowing the analysis to focus on the dynamics of semantic change rather than semantic content (see Supplementary Section S1 for more details on the method).

To characterize the stability of semantic progression across scales, we employ Allan deviation. For a time-indexed signal, Allan deviation measures how the difference between adjacent averages evolves with the averaging scale $\tau$, explicitly separating short-time fluctuations from long-time drift. In particular, a flattening of the Allan deviation curve indicates the emergence of a noise floor beyond which additional averaging fails to reduce variance. In the present context, this crossover defines a physically interpretable semantic stability horizon that cannot be identified using correlation-based methods alone.

In our analysis, the averaging scale $\tau$ denotes the number of sentences used in the coarse-graining of semantic distances. Short values of $\tau$ therefore probe sentence-to-sentence semantic variability, while larger $\tau$ reveal the accumulation of correlations over extended stretches of text. This interpretation allows semantic fluctuations to be analysed at different scales in a text corpus using similar methods prevalent in studying noise processes in physical systems. The results reported here do not depend on fine-tuned properties of a particular embedding model. While the primary analysis uses a sentence-transformer architecture (all-MiniLM-L6-v2), repeating the Allan deviation analysis across alternative transformer-based sentence embeddings yields quantitatively consistent short-time scaling exponents (see Supplementary Section S2 for multi-model comparison). Because Allan deviation probes only the scaling behaviour of cumulative increments, it is insensitive to absolute distances or embedding dimensionality. We therefore interpret the observed power-law regimes as reflecting ordered semantic progression in a text corpus rather than artifacts of representational choice.

As shown in Figure 1(c), at short scales, we observe power-law behaviour consistent with uncorrelated, white-noise-like semantic increments, a feature common across texts. At longer scales, Allan deviation may flatten, indicating the emergence of a noise floor beyond which large semantic shifts are not expected in the text corpus. This behaviour signals a breakdown of semantic convergence and motivates the definition of a characteristic context horizon.

For a stochastic signal exhibiting power-law scaling,
$$\sigma(\tau) \sim \tau^{\alpha},$$

the exponent $\alpha$ directly reflects the correlation structure of the underlying increments, with $\alpha = -1/2$ corresponding to uncorrelated noise and $\alpha \to 0$ indicating long-range correlations. In the following section, we quantify these short-time scaling exponents across genres and use them to compare different classes of text.

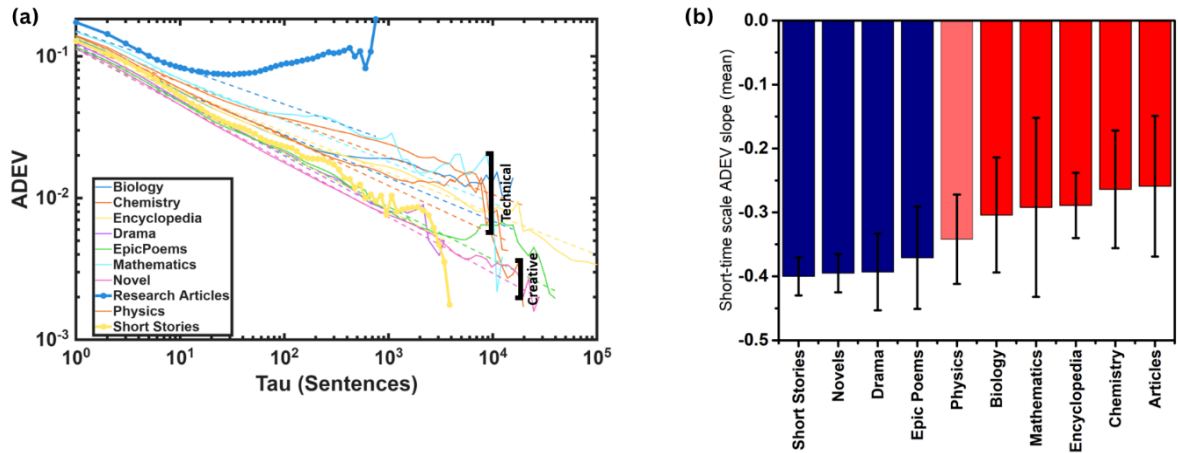

**Figure 2. Genre-dependent short-time scaling of semantic fluctuations.**
(a) Allan deviation of cumulative semantic phase for representative corpora spanning creative (novels, drama, epic poetry, short stories) and technical or informational genres (physics, mathematics, biology, chemistry, encyclopaedic text). Creative narratives exhibit steeper short-time scaling, clustering near the white-noise limit, whereas technical texts display flatter scaling indicative of stronger local correlations. (b) Mean short-time scale Allan deviation slope for each genre. Error bars indicate standard deviation across multiple texts from the same genre.

Figure 2(a) shows the Allan deviation $\sigma(\tau)$ of the semantic distances for representative creative and technical texts. We look at the exponent for a short-time regime (here restricted to $\lesssim 10\%$ of the maximum available scale) to probe local semantic organization rather than global narrative structure. At short averaging scales, all corpora exhibit clear power-law behaviour, $\sigma(\tau) \sim \tau^{\alpha}$, indicating scale-free fluctuations in semantic change. However, the scaling exponents separate by genre and we observe that creative literature including novels, short stories, drama, and epic poetry clusters near the white-noise limit, whereas technical and factual texts exhibit significantly shallower slopes as shown in Figure 2(b) and Table I.

Table I: Short time scale exponents averaged over genre

| Topic | Number of books | Short-time scale $\alpha$ (Mean±Std) |
|---|---|---|
| Biology | 20 | -0.304±0.09 |
| Chemistry | 17 | -0.264±0.092 |
| Encyclopedia | 11 | -0.289±0.051 |
| Drama | 37 | -0.393±0.06 |
| Epic Poems | 20 | -0.371±0.08 |
| Mathematics | 13 | -0.292±0.14 |
| Novels | 25 | -0.395±0.03 |
| Physics | 23 | -0.342±0.07 |
| PMC manuscripts | 412 | -0.259±0.11 |
| Short Stories | 25 | -0.4±0.03 |

Creative texts consistently produce short-time exponents closer to -0.4, indicative of weak correlations between successive semantic increments. In contrast, technical and informational texts exhibit significantly shallower exponents closer to -0.25. This reflects stronger local correlations and more constrained semantic evolution. Randomizing sentence order within each text eliminates these distinctions entirely (see Supplementary Section S3 for examples), producing behaviour consistent with a memoryless random walk. This confirms that the observed scaling regimes arise from ordered semantic progression rather than embedding geometry or text length.

The clear separation of scaling exponents across genres indicates that local semantic fluctuations encode structural information from sentence to sentence. This suggests that short-time Allan deviation exponents can serve as quantitative fingerprints of textual organization. Of particular interest is the physics corpus analysed here which consists primarily of expository and popular-science texts rather than technical research monographs. Such texts are explicitly designed to convey complex concepts through analogy, historical development, and narrative progression, often prioritizing conceptual continuity over formal derivation. We used this as a qualitative check to test that short-time scaling exponents are closer to those observed in narrative forms than in highly specialized scientific literature even for texts discussing highly technical concepts.

At larger averaging scales, many corpora exhibit a crossover from power-law behaviour at shorter time scales to a flattened Allan deviation regime. This signals the emergence of a noise floor where at such time scales large semantic shifts are not expected anymore. We define this crossover as a context horizon, identified as the smallest averaging scale at which the local Allan deviation slope deviates by more than 15% from its short-time value. Beyond this scale, additional averaging no longer reduces semantic variance, indicating a loss of effective semantic convergence. The location of the context horizon varies strongly across genres (Table II). Technical and informational texts exhibit crossovers at small fractions of their total length, whereas creative texts sustain power-law scaling of the Allan deviation over substantially larger scales. Notably, novels do not exhibit a comparable crossover within the measured range; instead, their short-time scaling persists across all accessible scales. Although effects of the finite size of our text collection cannot be excluded, this behaviour is consistent with scale-invariant semantic organization and contrasts sharply with technical texts, which rapidly develop a semantic noise floor.

Table II: Context Horizon

| Topic | Context Horizon (Normalized) |
|---|---|
| Biology | 4.125 |
| Chemistry | 5.34 |
| Encyclopedia | 3.785 |
| Drama | 45.88 |
| Epic Poems | 3.78 |
| Mathematics | 10.63 |
| Novels | No divergence >15%. Divergence >10% is observed at 11.58 |

| Physics | 27.38 |
| PMC manuscripts | 9.7 |
| Short Stories | 14.99 |

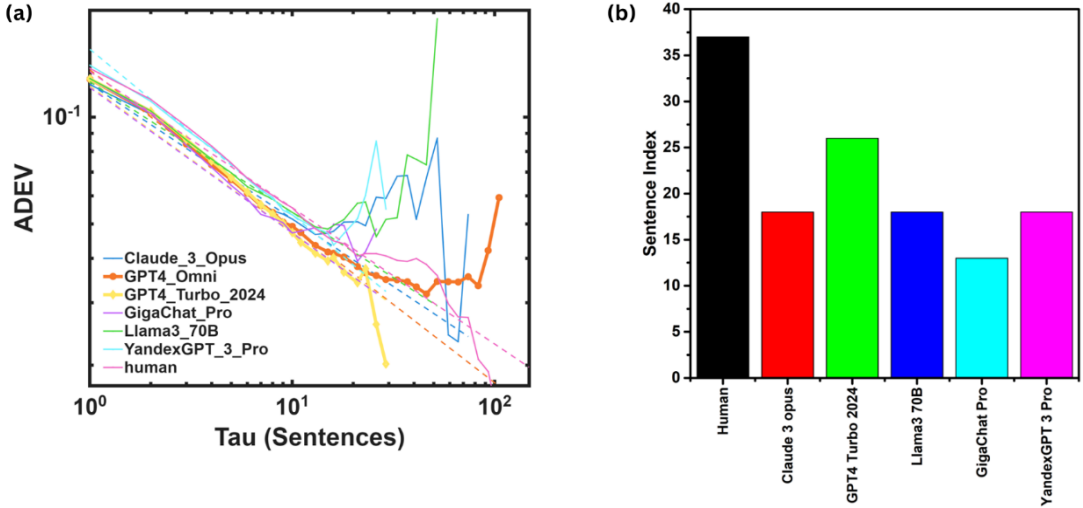

**Figure 3. Semantic stability in human and model-generated text.**
(a) Allan deviation of cumulative semantic phase for human-written text and multiple large language models generated under identical prompts. All systems exhibit comparable short-time power-law scaling, indicating statistically similar local semantic fluctuations. At larger averaging scales, model-generated text deviates earlier from the short-time regime. Dashed lines represent the short-time exponent.
(b) Context horizon, defined as the sentence index at which the Allan deviation slope deviates by more than 15% from its short-time value. Human-authored text sustains stable scaling over substantially larger sentence counts than all models, indicating a reduced semantic stability horizon for autoregressive generation.

The context horizon provides an operational scale for probing semantic drift[18], [19], [20], [21] in a text corpus. It marks the averaging scale beyond which additional context no longer contributes to emergence of new semantic paradigms in a text corpus. In this sense, the context horizon defines a stability-limited regime of semantic progression.

Figure 3(a) compares the Allan deviation of semantic phase trajectories for human-written and model-generated texts produced under identical prompts. At short averaging scales, all generative models exhibit power-law scaling with exponents comparable to those observed in human-authored text (Table III), indicating statistically similar local semantic fluctuations. We believe this captures the fact that model-generated texts are almost indistinguishable from human written texts at a sentence-to-sentence level.

Table III: Short-time exponent for human-written and generated texts.

| Writer | Sample Size | $\alpha$ (Mean±Std) |
| --- | --- | --- |
| Human | 802 | -0.384±0.15 |
| Claude 3 opus | 802 | -0.381±0.17 |
| GPT4 Turbo 2024 | 802 | -0.412±0.19 |
| Llama3 70B | 802 | -0.368±0.17 |
| GigaChat Pro | 802 | -0.409±0.19 |
| YandexGPT 3 Pro | 802 | -0.466±0.195 |

However, differences emerge at larger averaging scales when the analysis is performed as a function of absolute sentence index. For model-generated text, deviations from the short-time scaling regime occur at significantly smaller sentence counts than for human-written text. This behaviour is consistent across models and defines a reduced semantic stability horizon for current large language models (Table IV). In contrast, human-authored text maintains its short-time scaling behaviour over substantially longer sentence ranges under the same prompting conditions. The choice of absolute sentence index as the temporal variable is essential for isolating this effect. Large language models generate text autoregressively, such that semantic fluctuations accumulate sequentially. Sentence count therefore serves as a natural clock for generative stability, rather than by normalized text length. Normalization by total length of the output obscures this behaviour.

The earlier crossover observed in model-generated indicates that such texts often run out of semantically different ideas and converges to a core idea quickly leaving little room for exploration of a concept. We believe our method probes a fundamental shortcoming in generative models, since such models rely only on past meanings to generate the next token and are not designed to explore the semantic space to discover new connections.

Table IV: Context Horizon for human written and generated texts.

| Writer | Sample Size | Context Horizon (Sentence index) |
|---|---|---|
| Human | 802 | 37 |
| Claude 3 opus | 802 | 18 |
| GPT4 Turbo 2024 | 802 | 26 |
| Llama3 70B | 802 | 18 |
| GigaChat Pro | 802 | 13 |
| YandexGPT 3 Pro | 802 | 18 |

## III. Discussion

In this work, we introduced Allan deviation as a tool for quantifying the dynamics of semantic change in extended text. By mapping ordered sequences of sentence embeddings to a cumulative displacement signal, we treated narrative progression as a stochastic process indexed by sentence number. This approach reveals that semantic organization exhibits nontrivial scaling behaviours.

Across a wide range of corpora, we observed distinct short-time scaling regimes that separate creative literature from technical and factual writing. Creative texts cluster closer to the white-noise limit, indicating weak local correlations between successive semantic increments, whereas technical texts exhibit significantly stronger correlations. We believe this directly measures the freedom of semantic exploration inherent in creative literature and the constrained semantic space of factual writing. Creative texts permit relatively unconstrained local semantic variation while maintaining coherence through long-range structure, whereas technical and factual texts have stronger local dependencies to preserve conceptual consistency. Sentence-order randomization eliminates these distinctions, demonstrating that the observed scaling arises from ordered semantic structure rather than embedding geometry or text length.

At longer timescales, many corpora exhibit a crossover to a flattened Allan deviation regime, indicating the emergence of a noise floor beyond which averaging additional text does not reduce semantic variance. We formalized this crossover as a context horizon, defined operationally via deviations from short-time scaling. The location of this horizon varies systematically across genres. Applying the same analysis to large language models reveals a complementary distinction. While model-generated text reproduces human-like short-time scaling, deviations from this regime occur at smaller absolute sentence counts. Since generative models are not designed to explore the semantic space to discover new connections to divergent ideas, they quickly converge to the noise floor, when compared to human written texts.

These results establish Allan deviation as a physically interpretable tool for probing semantic dynamics across scale. The framework is agnostic to linguistic content, embedding choice, and model architecture, relying only on ordered semantic progression. It provides a bridge between statistical physics and language analysis, offering a quantitative route to study coherence, drift, and generative stability in both human and artificial text.

# Supplemental Information

## S1. Method of analysis

Data on genre specific books were manually downloaded from Project Gutenberg. Human and AI generated text dataset was taken from
https://huggingface.co/datasets/artnitolog/llm-generated-texts
PMC manuscripts were downloaded from
https://ftp.ncbi.nlm.nih.gov/pub/pmc/manuscript/txt/
An implementation of the methods described here is available here:
https://xortical.github.io/semanticrheology.html

### i. Sentence Embeddings and Semantic Increments

Texts are segmented into consecutive sentences, each mapped to a fixed-dimensional semantic embedding using a pretrained transformer model (all-MiniLM-L6-v2). Let $\mathbf{v}_t$ denote the embedding of sentence $t$. We define the instantaneous semantic increment as the angular (cosine) distance between successive embeddings,

$$\Delta\theta(t) = \arccos\left(\frac{\mathbf{v}_t \cdot \mathbf{v}_{t+1}}{\|\mathbf{v}_t\| \|\mathbf{v}_{t+1}\|}\right),$$

which isolates semantic change independently of vector norm.

### ii. Semantic Phase Construction

The cumulative semantic phase $\phi(t)$ is defined as the discrete sum of angular increments. $\phi(t) = \sum_{k=1}^{t} \Delta\theta(k)$. This transformation converts high-dimensional semantic motion into a scalar stochastic process while preserving temporal ordering. All analyses operate exclusively on $\phi(t)$; no information about absolute semantic location is retained.

### iii. Allan Deviation Formalism

For a discrete signal $\phi(t)$, the Allan deviation at averaging time $\tau$ is computed as

$$\sigma_y(\tau) = \sqrt{\frac{1}{2(N-2\tau)} \sum_{i=1}^{N-2\tau} [\phi(i+2\tau) - 2\phi(i+\tau) + \phi(i)]^2} \; \frac{1}{\tau}.$$

This definition emphasizes scale-dependent fluctuations and distinguishes correlated from uncorrelated stochastic processes. Averaging times $\tau$ are chosen on a logarithmic grid to ensure uniform coverage across scales.

### iv. Ensemble Averaging

Each text produces an individual Allan deviation curve. For each $\tau$, values are averaged across all texts contributing valid statistics at that scale. This avoids bias due to varying text lengths and ensures that long-$\tau$ behavior is dominated by sufficiently long texts.

### v. Scaling Exponent Extraction

Scaling exponents $\alpha$ are extracted by linear regression in log–log space,

$$\sigma_y(\tau) \sim \tau^\alpha,$$

performed separately for each text. Fits are restricted to $\tau \leq 0.1N$, where $N$ is the text length in sentences, to suppress finite-size roll-off. The ensemble mean and standard deviation of $\alpha$ quantify the robustness of the observed scaling.

### vi. Reference Curves and Visualization

Reference power-law curves shown in figures are constructed using the ensemble-averaged slope and anchored at a fixed intermediate scale. These curves are not fitted to the averaged Allan deviation but serve as visual guides derived independently from the slope statistics.

### vii. Robustness to Embedding Choice

Because the Allan deviation depends only on second differences of the cumulative phase, it is invariant under additive offsets, global rotations, and rescalings of the embedding space. Empirically, replacing the embedding model with alternative transformer-based architectures preserves the scaling exponents, confirming that the observed behavior reflects properties of semantic structure rather than embedding artifacts.

## S2. Model comparisons for short time scale slopes:

| Topic | Number of books | all-MiniLM-L6 (Mean±Std) | BGE-small slope (Mean±Std) | gte-small slope (Mean±Std) |
|---|---|---|---|---|
| Biology | 20 | -0.304±0.09 | -0.266±0.078 | -0.283±0.091 |
| Chemistry | 17 | -0.264±0.092 | -0.252±0.075 | -0.247±0.081 |
| Drama | 37 | -0.393±0.06 | -0.36±0.06 | -0.385±0.061 |
| Epic Poems | 20 | -0.371±0.08 | -0.345±0.077 | -0.356±0.088 |
| Mathematics | 13 | -0.292±0.14 | -0.258±0.13 | -0.286±0.125 |
| Novels | 25 | -0.395±0.03 | -0.375±0.033 | -0.384±0.026 |
| Physics | 23 | -0.342±0.07 | -0.29±0.049 | -0.299±0.058 |
| PMC manuscripts | 412 | -0.259±0.11 | -0.263±0.11 | -0.268±0.11 |
| Short Stories | 25 | -0.4±0.03 | -0.341±0.034 | -0.397±0.047 |

## S3. Null Test by breaking sentence orders:

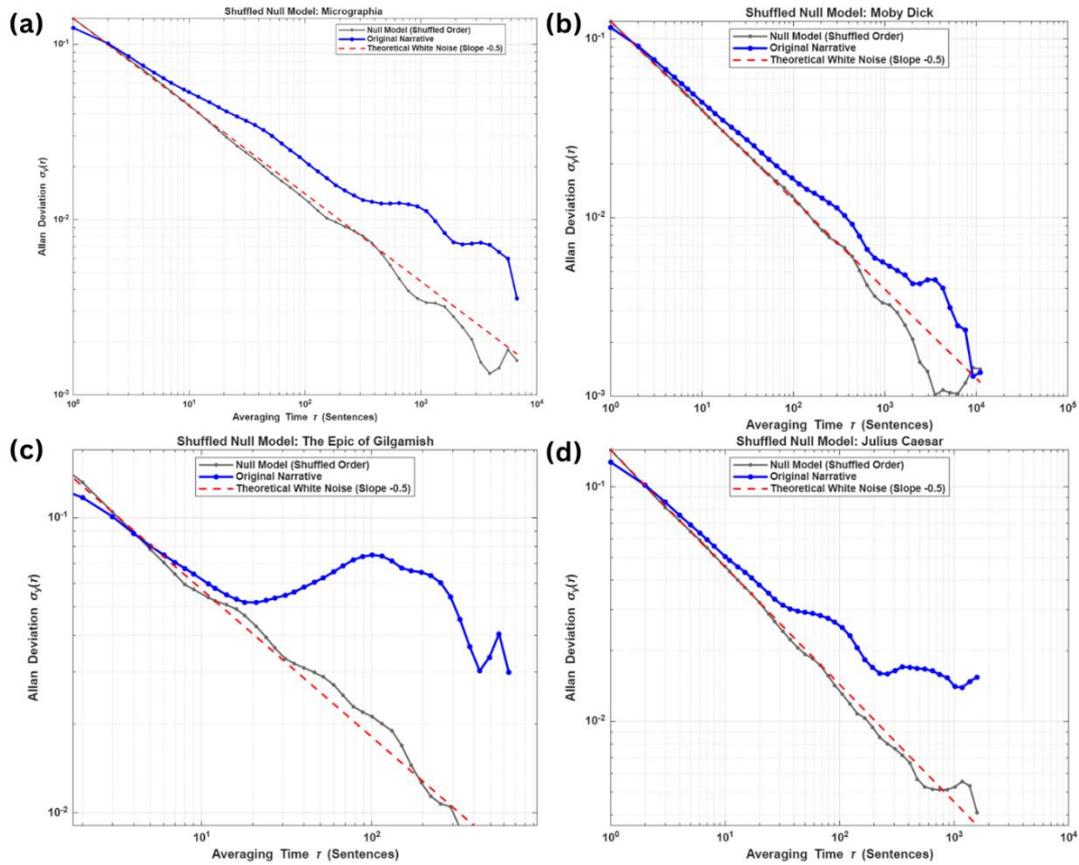

**Figure S1.** Allan deviation of cumulative semantic phase for original texts (blue) and sentence-order–randomized null models (black) for four representative corpora: (a) *Micrographia*, (b) *Moby Dick*, (c) *The Epic of Gilgamesh*, and (d) *Julius Caesar*. The red dashed line indicates the expected scaling for a memoryless random walk (white noise, slope −1/2). Original texts exhibit excess low-frequency variance and sustained deviations from white-noise behavior, reflecting long-range structure in ordered semantic progression. Sentence-order randomization suppresses this excess variance, causing the Allan deviation to collapse toward the white-noise reference across scales.

Randomizing sentence order removes the scale-dependent variance observed in coherent texts. After shuffling the sentence order and measuring the instantaneous and cumulative semantic displacements in the new text, the Allan deviation no longer exhibits persistent power-law scaling and instead approaches the behavior expected for a memoryless random walk. This demonstrates that the observed long-range structure is not a trivial consequence of embedding geometry or text length, but arises from the ordered sequence of semantic states. Consistent with this interpretation, the original texts exhibit excess low-frequency power that is suppressed by sentence-order randomization.